\documentclass[conference]{IEEEtran}

\usepackage{amsmath}
\usepackage{amsfonts}
\usepackage{graphicx}
\usepackage{cite}
\usepackage{multirow}
\usepackage{algorithm,algorithmic}
\usepackage{url}
\usepackage{enumerate}

\IEEEoverridecommandlockouts

\title{Deep Log-Likelihood Ratio Quantization\thanks{This work has been accepted for presentation at EUSIPCO 2019. Supported by grants NSF CNS 1731384, ARO W911NF-18-1-0247, and ONR N00014-19-1-2590.}}

\author{\IEEEauthorblockN{Marius Arvinte, Ahmed H. Tewfik and Sriram Vishwanath}
	\IEEEauthorblockA{Department of Electrical and Computer Engineering\\
		University of Texas at Austin\\
		Austin, Texas 78712\\
		Email: arvinte@utexas.edu}}
	
\begin{document}
	
	\maketitle
	
	\begin{abstract}
		In this work, a deep learning-based method for log-likelihood ratio (LLR) lossy compression and quantization is proposed, with emphasis on a single-input single-output uncorrelated fading communication setting. A deep autoencoder network is trained to compress, quantize and reconstruct the bit log-likelihood ratios corresponding to a single transmitted symbol. Specifically, the encoder maps to a latent space with dimension equal to the number of sufficient statistics required to recover the inputs --- equal to three in this case --- while the decoder aims to reconstruct a noisy version of the latent representation with the purpose of modeling quantization effects in a differentiable way. Simulation results show that, when applied to a standard rate-1/2 low-density parity-check (LDPC) code, a finite precision compression factor of nearly three times is achieved when storing an entire codeword, with an incurred loss of performance lower than 0.1 dB compared to straightforward scalar quantization of the log-likelihood ratios.
	\end{abstract}
	
	\section{Introduction}
	\label{section_intro}
	
	Quantization is a key component of communication networks and has been intensely researched in the past years. Since practical systems are often memory-limited, quantizing either channel observations or derived information, such as log-likelihood ratios, using as few bits as possible is of critical importance. At the same time, considerable research in deep learning and data science broadly aims to answer the question: \textit{Given high-dimensional data, is there a low-dimensional representation of it}? Considering the log-likelihood ratios derived from transmitting over a memoryless channel, the answer is an immediate \textit{yes}, since, given the channel observation and state information, their values can be derived exactly.
	
	In this paper, we propose a deep learning-based LLR quantization algorithm that empirically answers the question: \textit{Given a set of log-likelihood ratios, is there a low-dimensional representation of it that is robust to quantization}? Our results show that we are able to construct such a highly nonlinear representation by using a deep autoencoder.
	
	Quantization with one-bit analog-to-digital converters has been extensively studied, such as in \cite{one_bit_heath}, where the authors present theoretical results regarding the capacity, revealing that the regime of one-bit quantization is benefited by a low operating signal-to-noise ratio (SNR) and increased number of receive antennas. In \cite{kurkoski}, the effects of quantizing the output of a binary-input discrete memoryless channel are studied and the optimal quantizer for a given number of levels is derived.

	The work in \cite{mutual_inf_llr} investigates optimal scalar LLR quantization and designs a quantizater that aims to maximize mutual information, while in \cite{vq_llr_mi} the authors design a vector quantizer with the same objective and investigate its performance in a Turbo-coded, hybrid automatic repeat request (HARQ) scenario, demonstrating that an equivalent precision of three bits per value is sufficient to retain most of the performance.
	
	On the deep learning side, the past years have seen a resurgence of deep neural networks (DNN) as \textit{universal function approximators} \cite{universal_nn_approx}, as well as a very rapidly increasing interest in using deep learning to solve problems in communication systems \cite{survey_dl_comm}. One particularly appealing architecture for communication problems is the \textit{deep autoencoder} \cite{ae_basic}. Broadly speaking, an autoencoder aims to approximate the function $f(x) = x$ under certain structural constraints placed on its \textit{latent representation}. Typically, these constraints can be related either to the dimension of the latent space or to the signal structure itself.
	
	Previous work in \cite{oshea_learning_comm} uses an autoencoder to learn novel, highly nonlinear modulation schemes that offer an implicit coding gain and are robust to channel variations. In another example, the authors of \cite{ae_ofdm} propose an OFDM-autoencoder structure that performs a pair of time/frequency transforms in the latent domain to impose structure on it similar to that of an OFDM signal. Using autoencoders for compression has been extensively studied in computer vision, where they are used for lossy compression of images \cite{ae_img_compression}, with results showing that they are competitive with state-of-the-art JPEG compression, as well as being used for image denoising.
	
	Motivated by these recent results, we propose the use of a deep autoencoder for log-likelihood ratio quantization and storage at the receiver side, in a single-input single-output uncorrelated fading channel scenario. We recognize that a latent space of dimension three captures all sufficient statistics and use this in designing the autoencoder. There are three key components to the proposed architecture:
	
	\begin{enumerate}[i)]
		\item A loss function that is weighted towards more accurate reconstruction of low (uncertain) LLR values.
		\item A latent representation of dimension equal to the number of sufficient statistics required to reconstruct the LLR values for each channel use.
		\item A Gaussian noise layer that approximates numerical quantization noise.
	\end{enumerate}

	During inference, the receiver uses the encoder to compress the vector of LLR values corresponding to a modulated symbol to its latent representation, performs simple uniform scalar quantization and stores the values for future use. The decoder, or a predetermined lookup table, is used to reconstruct the LLR values from this quantized latent representation.
	
	 The performance of our proposed method is evaluated when coupled with high-order quadrature amplitude modulation (QAM) and LDPC coding in both standalone and HARQ scenarios, where we show that a compression factor of almost three times is achieved with minimal loss of performance, ultimately enabling more efficient buffer usage for memory-limited applications.
	
	\section{Preliminaries}
	\label{section_sys}

	\subsection{System Model}
	Let $s$ be the transmitted symbol drawn from a constellation $\mathcal{C}$ with cardinality $2^K$, corresponding to bits $b_i, i=1,\dots,K$. Then, the received symbol $r$, after passing through a single-input single-output memoryless fading channel, is given by
	
	\begin{equation}
		r = hs + n,
	\end{equation}
	
	\noindent where $n \sim \mathcal{CN}(0, \sigma_n^2)$ is the complex Gaussian noise and $h$ is the i.i.d. complex channel coefficient.
	
	Assuming equal prior probabilities for all bits and that the receiver has complete channel state information, the log-likelihood ratio of bit $b_i$ is given by \cite{gallager_book}
	
	\begin{equation}
	\label{eq_llr_exact}
		L_i = \frac{P(r | b_i = 1)} {P(r | b_i = 0)} =
		 \frac {\sum\limits_{s \in \mathcal{C}, b_i = 1} \exp{-\frac{\lvert r - hs \rvert^2}{\sigma_n^2}}}
		 {\sum\limits_{s \in \mathcal{C}, b_i = 0} \exp{-\frac{\lvert r - hs \rvert^2}{\sigma_n^2}}}.
	\end{equation}
	
	Factoring out the $h$ term and letting $\tilde{r} = r/h$ yields
	
	\begin{equation}
	L_i = \frac {\sum\limits_{s \in \mathcal{C}, b_i = 1} \exp{-\frac{\lvert h\rvert^2}{\sigma_n^2}} \lvert \tilde{r} - s \rvert^2}
	{\sum\limits_{s \in \mathcal{C}, b_i = 0} \exp{-\frac{\lvert h\rvert^2}{\sigma_n^2}} \lvert \tilde{r} - s \rvert^2}.
	\end{equation}
	
	Let
	
	\begin{equation}
	\label{eq_stats}
		G = \Big( \frac{\lvert h \rvert}{\sigma_n} \Big) ^2, \tilde{r}_r = \Re\{\tilde{r}\}, \tilde{r}_i = \Im\{\tilde{r}\},
	\end{equation}
	
	\noindent then the real-valued vector $(G, \tilde{r}_r, \tilde{r}_i) \in \mathbb{R}^3$ is a sufficient statistic to determine the log-likelihood ratios $L_i$ corresponding to a single channel use \cite{gallager_book}. Note that $G$ is exactly the instantaneous SNR.
	
	Thus, as long as we have access to, or exactly recover, any bijective function of the vector of real-valued sufficient statistics, the exact log-likelihood ratios can be computed as well. Let
	
	\begin{equation}
	\label{eq_soft}
		\Lambda_i = \tanh \Big( \frac{L_i}{2}\Big),
	\end{equation}
	
	\noindent be the \textit{soft bit} associated to the $i$-th bit \cite{mutual_inf_llr}, with the immediate property that $\Lambda_i \in [-1, 1]$. This will prove useful in training our deep neural network, since large values are implicitly compressed by this transformation.
	
	\subsection{Deep Autoencoders}
	Broadly speaking, an autoencoder is a multilayer neural network that aims to approximate the identity function between its inputs and outputs under specific constraints placed on an intermediate output termed \textit{latent representation}.
	
	Letting $\mathbf{x}$ be the input to the autoencoder, $\mathbf{y}$ its output and $\mathbf{z}$ the latent representation, then the input-output relationship is given by
	
	\begin{equation}
		\mathbf{y} = g\big(\mathbf{z}; \mathbf{\theta}_g) = g(f(\mathbf{x}; \mathbf{\theta}_f); \mathbf{\theta}_g\big),
	\end{equation}
	
	\noindent where $f$ and $g$ are the \textit{encoder} and \textit{decoder}, parametrized by weights $\theta_f$ and $\theta_g$, respectively. Abusing notation and dropping the weights, the loss function of a general autoencoder, given a dataset $(\mathbf{X})_i, i = 1, \dots, N$, can be expressed as
	
	\begin{equation}
		\mathcal{L}(\mathbf{X}) = \sum\limits_{i=1}^N \mathcal{D} \big( \mathbf{x}_i, g(f(\mathbf{x}_i)) \big),
	\end{equation}
	
	\noindent where $\mathcal{D}$ is a differentiable distance function.
	
	Typical constraints placed on the latent representation include that it is restricted to a low dimensional subspace e.g. $\mathbf{z} \in \mathbb{R}^M$, with $M \ll \dim(\mathbf{x})$, or that it is noised during training, for each sample, to yield
	
	\begin{equation}
		\tilde{\mathbf{z}}_i = \mathbf{z}_i + \mathbf{n}_i,
	\end{equation}
	
	\noindent where $\mathbf{n}$ is typically i.i.d. Gaussian with adjustable mean and variance \cite{ae_img_compression}. Note that this operation is differentiable and has unit gradient, thus can be used in conjuction with backpropagation during training.
	
	\section{The Proposed Scheme}
	\label{section_prop_alg}
	The deep LLR quantization scheme consists of two parts:
	
	\begin{enumerate}[i)]
		\item An autoencoder with a weighted loss function, a latent space of dimension three and a Gaussian noise layer active during training.
		\item A scalar quantization function.
	\end{enumerate}

	Assuming the soft bits $\Lambda_i$ are already computed for the current symbol using (\ref{eq_soft}), the goal is to store them in finite precision using as few bits as possible, with a performance degradation as small as possible. Since soft bits with low absolute values are more prone to causing error propagation when perturbed in modern decoding algorithms (e.g., \cite{ldpc_errors}), we propose to use as distance function the weighted squared loss
	
	\begin{equation}
		\mathcal{D}(\Lambda_i, \tilde{\Lambda}_i) = \sum\limits_{j=1}^K \frac{\lvert \Lambda_j - \tilde{\Lambda}_j \rvert^2}{\lvert \Lambda_j \rvert + \epsilon},
	\end{equation}
	
	\noindent where $\epsilon = 10^{-4}$ prevents division by very small amounts and numerical degeneration. Thus, the loss function of the autoencoder becomes
	
	\begin{equation}
		\mathcal{L}(\mathbf{\Lambda}, \tilde{\mathbf{\Lambda}}) = \sum\limits_{i=1}^N \sum\limits_{j=1}^K \frac{\lvert \Lambda_j^{(i)} - \tilde{\Lambda}_j^{(i)} \rvert^2}{\lvert \Lambda_j^{(i)} \rvert + \epsilon},
	\end{equation}
	
	\noindent where $\Lambda_j^{(i)}$ is the $j$-th soft bit of the $i$-th training sample and $\tilde{\Lambda}_j^{(i)} = g\big(f(\Lambda^{(i)})\big)_j$ is its autoencoder reconstruction.
	
	This enables the autoencoder to learn more accurate representations of soft bits with low absolute values, and ultimately log-likelihood ratios with the same property, since $\tanh$ is approximately linear around zero. At the same time, emphasis is placed on reconstructing soft bits with absolute value close to $1$, where $\mathcal{D}$ becomes closer to the squared distance.
	
	The dimension of the latent representation is constrained to three, since this is the number of sufficient statistics for recovering the log-likelihood ratios, as given by (\ref{eq_stats}). Using an additive Gaussian noise layer with zero mean and variance $\sigma_t^2$, the autoencoder is trained to learn a robust latent representation of the soft bits. While, ideally, we would like to incorporate quantization directly in the training procedure, this is not possible in a straightforward manner because the step function has zero differential almost everywhere, hence is not compatible with modern gradient descent training methods. In this sense, we effectively approximate the quantization error with Gaussian noise.
	
	The last layer of the encoder and the decoder are designed to use $\tanh$ as activation function. For the decoder, it is straightforward to do so, since it helps constrain the outputs in the $[-1, 1]$ interval, where the input signal is found. For the encoder, the reasoning behind this is twofold:
	
	\begin{enumerate}[i)]
		\item $\tanh$ serves as an implicit regularization counter to adding Gaussian noise. If the output of the encoder was unbounded (e.g. linear or ReLU activation), then the autoencoder would learn a robust $\mathbf{z}$ by simply increasing its energy indefinitely.
		
		\item Constraining $\mathbf{z}$ to lie in $[-1, 1]^3$ makes it more amenable to subsequent finite precision quantization, allowing the use of the same quantizer for all components $z_i$.
	\end{enumerate}

	Once training is complete, the soft bit vector $\mathbf{\Lambda}$ corresponding to a symbol is input to the encoder, resulting in the latent representation $\mathbf{z} = (z_1, z_2, z_3)$. $\mathbf{z}$ is elementwise quantized to finite precision using the function
	
	\begin{equation}
		Q(x) = \left\{ \begin{array}{ll}
		\text{sgn}\{x\}\delta & \text{if} \ \lvert x \rvert > \delta \\
		\frac{\delta}{2^{N_b}} \Delta_k & \text{if} \ x \in -\delta + [\frac{k\delta}{2^{N_b-1}}, \frac{(k+1)\delta}{2^{N_b-1}}] \\
		\end{array}
		\right. ,
	\end{equation}
	
	\noindent where $\Delta_k = -2^{N_b} + 2k + 1$.	Thus, all values outside the range $[-\delta, \delta]$ are first saturated to $\text{sgn}\{x\}\delta$ and subsequently uniformly quantized using $N_b$ bits.
	
	Reconstructing the soft bits from the latent representation $\mathbf{z}$ is done by simply passing it through the decoder $g$. Thus, the reconstructed soft bits, as a function of the original values, take the form
	
	\begin{equation}
		\tilde{\mathbf{\Lambda}} = g\big(Q(f(\mathbf{\Lambda}))\big).
	\end{equation}
	
\begin{figure}[!t]
		\centering
		\includegraphics[width=3.4in]{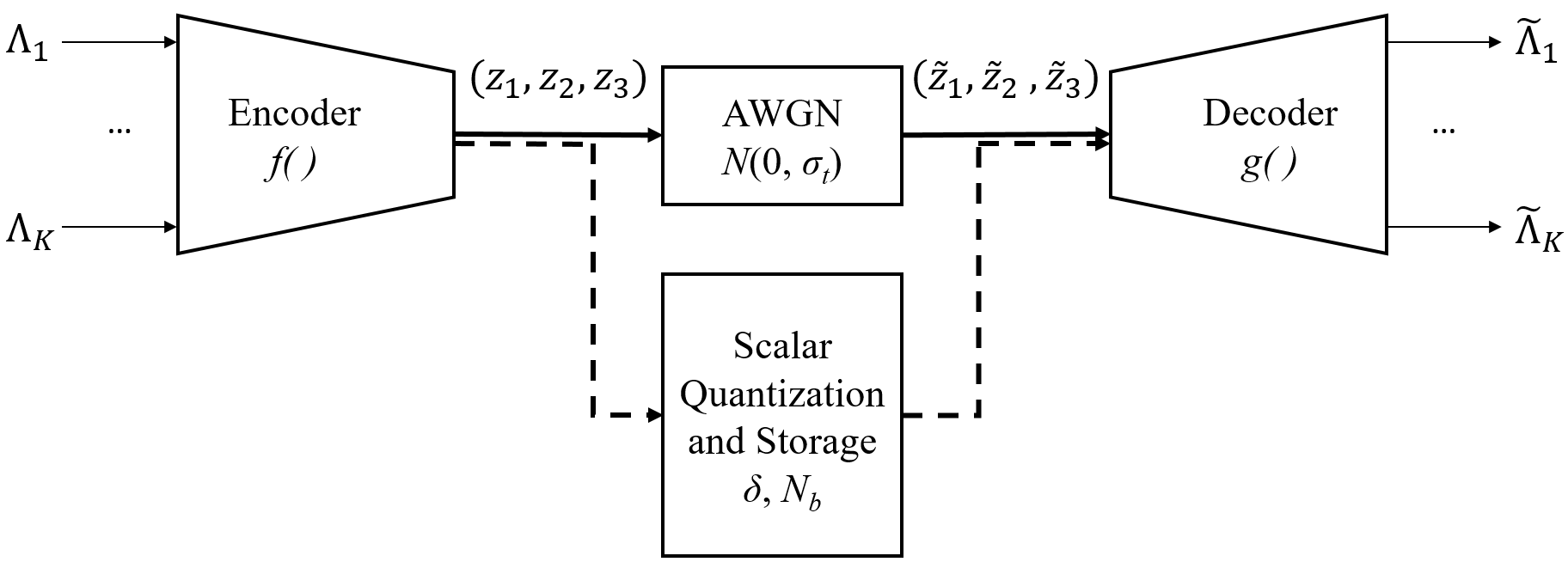}
		\caption{The architecture of the proposed scheme. Solid lines correspond to the signal path during training, where $\tilde{\mathbf{z}}$ represents the noisy latent representation of the input soft bits. Dashed lines correspond to the inference path, where the same notation represents the quantized latent representation of $\mathbf{z}$, which is stored for future use.}
		\label{fig_block_diagram}
	\end{figure}

	\figurename{ \ref{fig_block_diagram}} shows the architecture of the proposed scheme, both during training and inference time. We use a symmetrical architecture for the autoencoder, in which the encoder and decoder have the same number of hidden layers, with their output dimensions mirrored.
	
	\section{Performance Results}
	\label{section_perf}
	
	The performance of our proposed algorithm is simulated when used together with a rate-1/2 LDPC code of codeword length $N=648$ and 256-QAM modulation, used in the WiFi 802.11 family of standards. Additionlly, we discuss how the autoencoder achieves robustness to quantization in the latent domain.
	
	\begin{figure}[!t]
		\centering
		\includegraphics[width=2.4in]{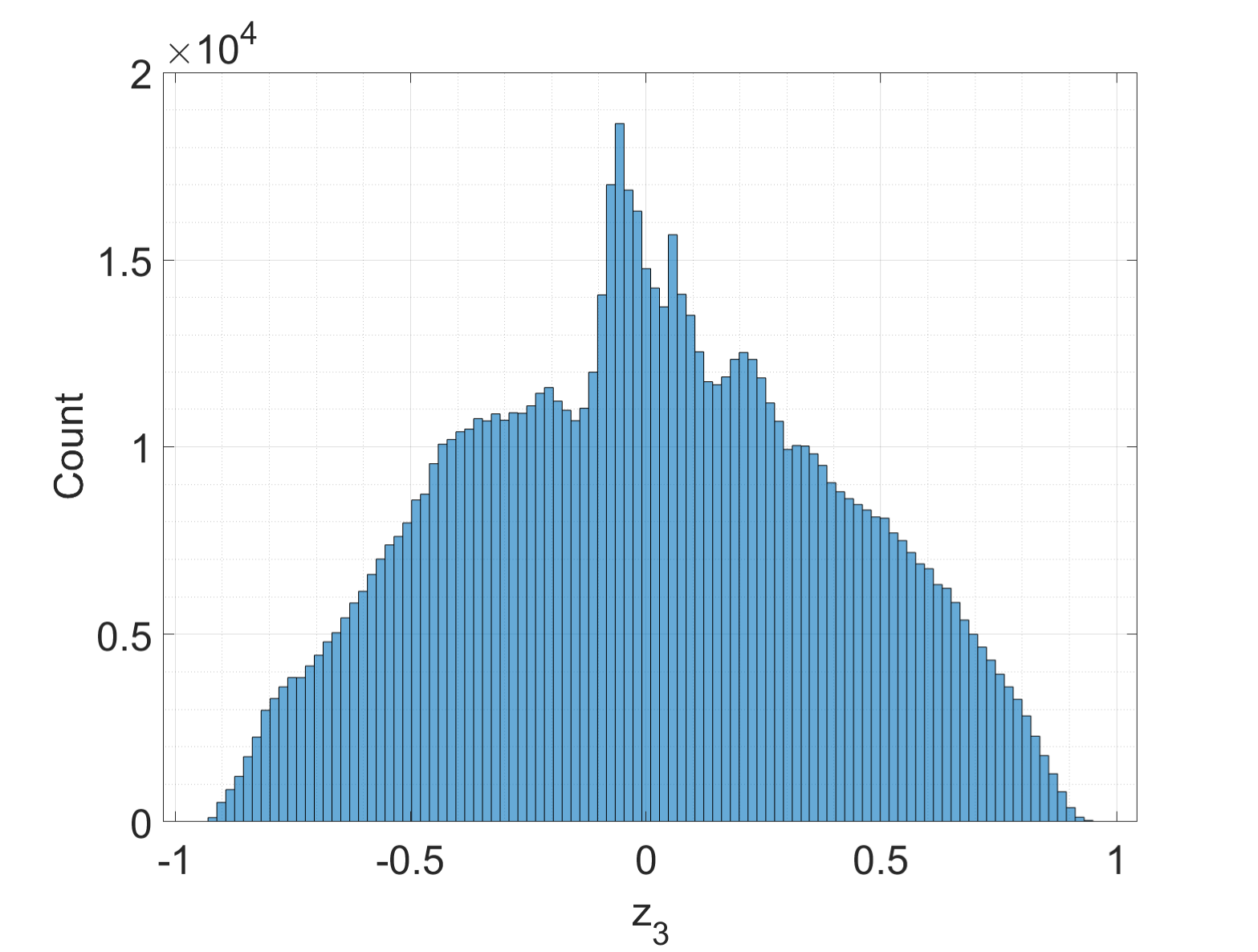}
		\caption{Empirical distribution of $z_3$ at the SNR value of 18 dB, 256-QAM. Obtained by encoding the LLR values from $10000$ LDPC codewords.}
		\label{fig_z3_hist}
	\end{figure}
	
	To construct training and validation datasets, we generate random bit payloads, encode them, interleave and modulate the resulted codeword and apply an i.i.d. Rayleigh fading channel (i.e., $h \sim \mathcal{CN}(0,1)$) for each QAM symbol. We compute the exact expressions of the log-likelihood ratios by using (\ref{eq_llr_exact}) and use them as training data. We do this for multiple SNR values, generating $10000$ codewords for each one, and concatenate the resulted values together, allowing the proposed architecture to work without requiring separate SNR estimates as input.

	Once the autoencoder is trained, we design our latent space uniform quantizer. \figurename{ \ref{fig_z3_hist}} plots the marginal distribution of the latent signal $z_3$ when LLR values at a specific SNR are encoded. Empirically, we choose $\delta = 0.8$ as clipping threshold for all latent signals. Details of the training and validation procedures are given in Table \ref{table_training}. The simulations were performed using the Keras library \cite{keras_cite}, as well as Matlab. The source code is available online\footnote{\url{https://github.com/mariusarvinte/deep-llr-quantization}}, alongside extended simulation results.

	\begin{table*}[!t]
		\renewcommand{\arraystretch}{1.3}
		\caption{Architectural and Training Details of the used Autoencoder}
		\label{table_training}
		\centering
		\begin{tabular}{|c|c|c|}
			\hline
			\textbf{Hidden layers} & \textbf{Output size} & \textbf{Activation function} \\
			\hline
			Feedforward, fully-connected & Intermediate layers: $4K$ & ReLU (except last layer of $f$/$g$) \\
			6 total, 3 per $f$/$g$ & Latent representation: $3$ & $\tanh$ \\
			\hline
			\textbf{Noise layer} & \textbf{Training algorithm} & \textbf{Quantizer} \\
			\hline
			$\mu_t = 0, \sigma_t^2 = 10^{-6}$ & Adadelta, $\text{LR} = 2, \text{Decay} = 0.8$ & $\delta = 0.8$, Uniform \\
			\hline
		\end{tabular}
	\end{table*}
	
	\subsection{Single Transmission Performance}	
	The packet error rate performance is simulated in a single transmission scenario. Since prior work on LLR quantization shows that a number of 4-5 bits usually retains most of the decoding performance \cite{mutual_inf_llr}, our architecture is capable of achieving a higher compression ratio as we target more densely packed constellations.
	
	For the reference quantization scheme, we also perform saturation outside the $[-4, 4]$ interval, empirically determined to provide a good tradeoff. We also simulate the performance of applying scalar quantization to the vector $(G, \tilde{r}_r, \tilde{r}_i)$ to highlight the very large error floor it creates when limiting the number of bits. In particular, we know that $G$ has a Rayleigh distribution for our channel model, thus we apply the optimal scalar quantizer in \cite{rayleigh_old_q} for it.
	
\begin{figure}[!t]
		\centering
		\includegraphics[width=3.6in]{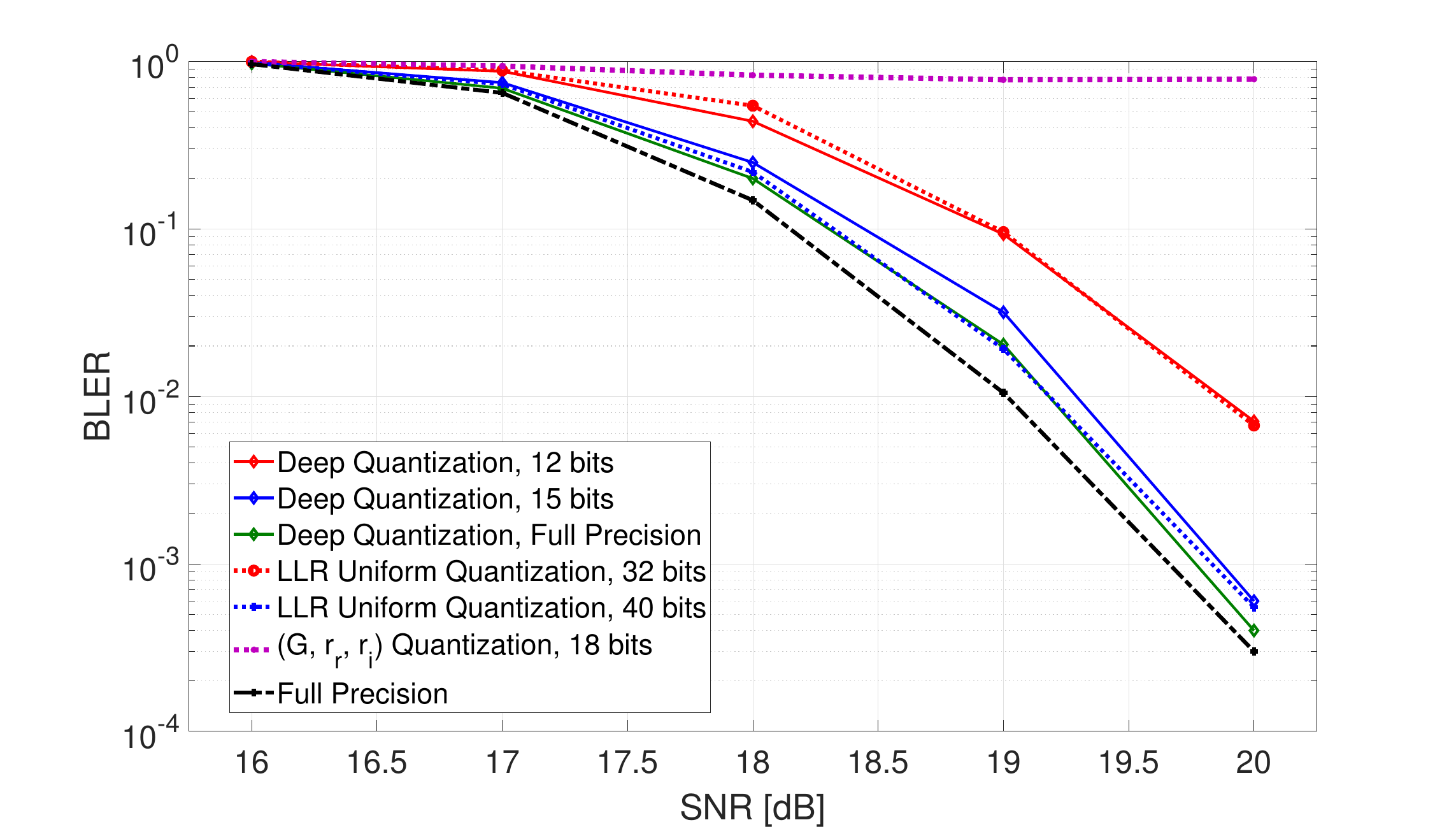}
		\caption{Block error rate performance for the single transmissions scenario, using 256-QAM modulation. For each SNR value, $10000$ LDPC codewords are simulated, separate from the training data used by the deep quantization method.}
		\label{fig_plain_ldpc}
	\end{figure}

	\figurename{ \ref{fig_plain_ldpc}} shows the performance results obtained. Comparing at $10^{-2}$ block error rate, we notice there is a performance loss of 0.15 dB incurred by the autoencoder even without any quantization of the latent representation. This represents a fundamental tradeoff between robustness to quantization and reconstruction precision.
		
	Using the data in \figurename{ \ref{fig_z3_hist}}, the mean and variance of $z_3$ are estimated to be approximately 0 and 0.158, respectively. Coupled with the injected noise power of $\sigma_t^2 = 10^{-6}$, the effective SNR in the latent space is approximately 52 dB. This appears large compared to the actual noise introduced by quantizing to 3-5 bits, since it corresponds to an effective resolution of 8 bits, but we find that, if $\sigma_t$ is further increased, the autoencoder does not converge to a solution with low reconstruction error, since too much noise is added to the latent representation.
	
	\figurename{ \ref{fig_plain_ldpc}} shows that by quantization of the latent representation and subsequent reconstruction, the performance of the reference scheme is retained by storing a total of 12 vs. 32 bits per symbol or 15 vs. 40 bits with a performance degradation lower than 0.1 dB, thus achieving a compression factor of approximately $2.7$ times.
	
	\subsection{Multiple Transmission Performance}
	The same autoencoder and channel code are used in a HARQ scenario with two transmissions, each sending a random, but fixed, half of the codeword and an additional one third of bits from the other half, having an effective code rate of $3/8$. The log-likelihood ratios of the first transmission are quantized, while the ones from the second transmission are assumed to be available in full precision. The log-likelihood ratios common to both transmissions are combined using equal gain combining, given by the expression
	
	\begin{equation}
		L_i = \tilde{L}_i^{(1)} + L_i^{(2)},
	\end{equation}
	
	\noindent where $\tilde{L}_i^{(1)}$ is the reconstructed version of the quantized ratio $L_i^{(1)}$ and $L_i^{(2)}$ is the full precision LLR from the second transmission.
	
	\begin{figure}[!t]
		\centering
		\includegraphics[width=3.6in]{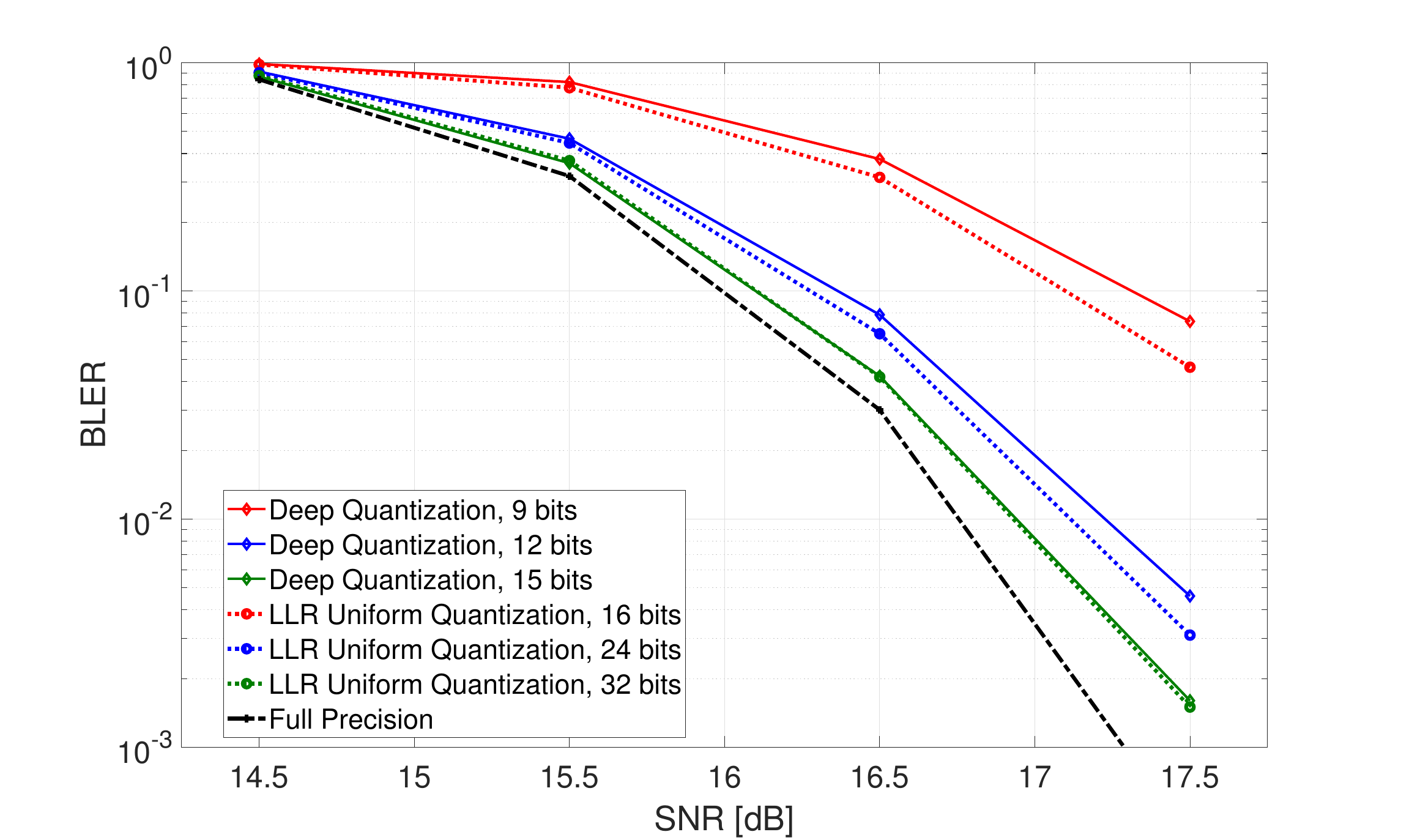}
		\caption{Block error rate performance for the multiple, overlapping transmissions scenario, using 256-QAM modulation. For each SNR value, $10000$ LDPC codewords are simulated, separate from the training data used by the deep quantization method.}
		\label{fig_harq_ldpc}
	\end{figure}
	
	\figurename{ \ref{fig_harq_ldpc}} shows the performance results obtained with a 256-QAM modulation, where we notice that the proposed 15 bit representation has near identical performance to 32 bits in the reference implementation, achieving a compression ratio of $2.13$ times, at a performance very close to full precision. Alternatively, if a more significant loss of performance versus the full precision is tolerated, deep quantization can be used with 9 bits and a compression ratio of $1.77$ times, approaching the hard-output decoding limit of one equivalent bit per LLR value.
	
	\subsection{Discussion}
	We now provide an interpretation of what the autoencoder learns for its representation. Since the latent space dimension is constrained to the number of sufficient statistics that describe the log-likelihood ratios, it is reasonable to assume that any valid (i.e., with reconstruction error sufficiently small) representation is approximately a bijective, potentially highly nonlinear, function of the statistics
	
	\begin{equation}
		(z_1, z_2, z_3) \approx \Phi(G, \tilde{r}_r, \tilde{r}_i).
	\end{equation}
	
	Training with added Gaussian noise in the latent space drives the autoencoder to learn a function $\Phi$ of the sufficient statistics that is robust to quantization noise, assuming the approximation is reasonable. Note that the identity mapping is not a suitable function in this case, since quantizing $G = \lvert h \rvert^2 / \sigma_n^2$, in particular, introduces a severe error floor, as seen in \figurename{ \ref{fig_plain_ldpc}}. 
	
\begin{figure}[!t]
		\centering
		\includegraphics[width=2.4in]{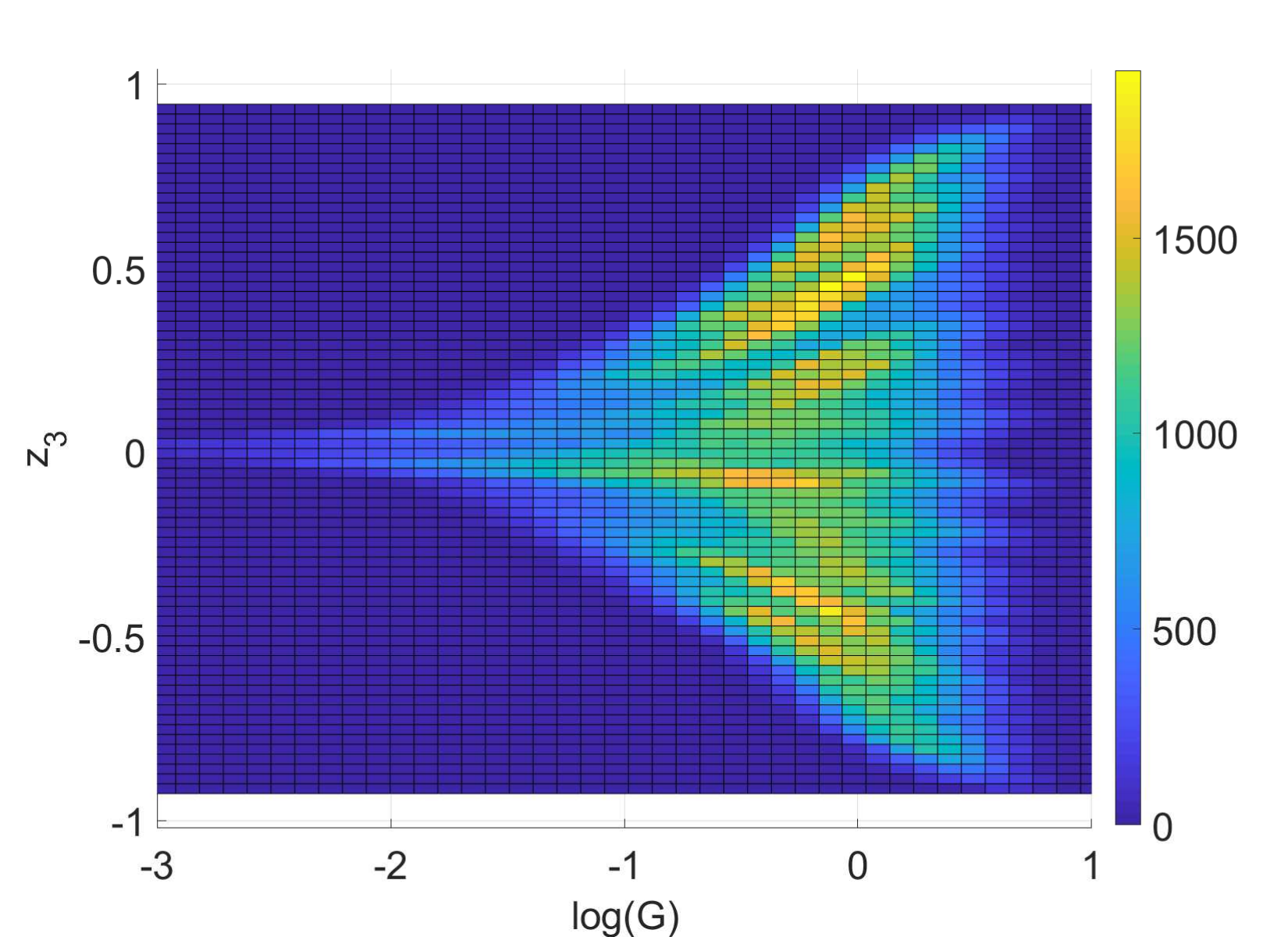}
		\caption{Empirical joint distribution of $(\log{G}, z_3)$ in the same conditions as \figurename{ \ref{fig_z3_hist}}. The intensity of the color is proportional to the probability mass of the respective bin.}
		\label{fig_G_z3_hist}
	\end{figure}

	Thus, the learned mapping $\Phi$ is potentially a highly non-linear function. This is indeed the case when visualizing the empirical joint distribution between each sufficient statistic and each latent signal component. \figurename{ \ref{fig_G_z3_hist}} plots the joint distribution of $\log(G)$ and $z_3$, where it is observed that this is not a one-to-one mapping, thus $z_3$ contains information on at least one more sufficient statistic. While this provides an interpretation of what the autoencoder learns for a Rayleigh fading channel, the structure is fragile to changing the distribution of $h$, where retraining the entire structure is required. This can be overcome by applying techniques from the field of machine learning under covariate shift, which is left for future work.
	
	Finally, note that since the latent space is quantized in a finite number of points, this is also true for the reconstructed log-likelihood ratios. Thus, if desired, one can precompute all possible reconstructions, store them in a larger, external memory and eliminate the decoder from the inference stage.
	
	\section{Conclusions}
	\label{section_conc}
	
	In this paper, a deep learning-based log-likelihood ratio quantization algorithm for uncorrelated fading channels was introduced. It is recognized that the number of sufficient statistics required to recover the LLR values is equal to three and an autoencoder is designed and trained to extract a latent representation of that exact dimension, while being robust to quantization noise when recovering the LLR values.
	
	Using an autoencoder and a uniform quantization function, we have shown that we can achieve compression factors of $2.7$ and $2$ times for standalone and HARQ scenarios, respectively, when a standard rate-1/2 LDPC code is used in conjuction with 256-QAM modulation. Other advantages are a width of the hidden layers on the order $\mathcal{O}(K)$ and the fact that the same autoencoder is trained and used across a wide range of SNR values. The proposed scheme has a wide range of applications, including, but not limited to, HARQ scenarios, relay and feedback channels.
	
	\bibliographystyle{IEEEtran}
	\bibliography{myBib}
	
\end{document}